%% file: main.tex
\documentclass[10pt,twocolumn,letterpaper]{article}

\usepackage{cvpr}              %

\usepackage{graphicx}
\usepackage{amsmath}
\usepackage{amssymb}
\usepackage{booktabs}
\usepackage{_mathalias}

\input{preamble}

\def\paperID{7590} %
\def\confName{CVPR}
\def\confYear{2025}

\begin{document}
\title{3D-AVS: LiDAR-based 3D Auto-Vocabulary Segmentation}

\author{
Weijie Wei$^*$,\hspace{0.4em} Osman Ülger$^*$,\hspace{0.4em} Fatemeh Karimi Nejadasl,\hspace{0.4em} Theo Gevers,\hspace{0.4em} Martin R. Oswald \\
University of Amsterdam, the Netherlands\\
}

\maketitle
\blfootnote{$^*$ Co-first author}

\input{sec/0_abstract}    
\input{sec/1_intro}

\input{sec/2_related_work}
\input{sec/3_methodology}
\input{sec/4_evaluation}
\input{sec/5_experiments}

\input{sec/6_conclusion}

\input{sec/X0_acknowledgement}

{
    \small
    \bibliographystyle{ieeenat_fullname}
    \bibliography{egbib}
}

\clearpage
\beginsupplement
\input{sec/X_suppl}

\end{document}

%% file: preamble.tex
\usepackage{colortbl}
\usepackage{graphicx}
\usepackage{amsmath}
\usepackage{amssymb}
\usepackage{booktabs}
\usepackage{multirow}
\usepackage{makecell}
\usepackage{enumitem} %
\usepackage{soul}
\usepackage{tabularx}

\usepackage[accsupp]{axessibility}
\usepackage{mmstyle}

\usepackage{nomencl} 
\usepackage{etoolbox} %

\definecolor{MyRed}{HTML}{FF0000}

\newcommand{\tablestyle}[2]{\setlength{\tabcolsep}{#1}\renewcommand{\arraystretch}{#2}\centering\small}
\newcommand{\blfootnote}[1]{%
  \begingroup
  \renewcommand\thefootnote{}%
  \footnote{#1}%
  \addtocounter{footnote}{-1}%
  \endgroup
}

\usepackage[misc]{ifsym}

\usepackage{pifont}

\definecolor{cvprblue}{rgb}{0.21,0.49,0.74}
\usepackage[pagebackref,breaklinks,colorlinks,allcolors=cvprblue]{hyperref}

\usepackage[capitalize]{cleveref}
\crefname{section}{Sec.}{Secs.}
\Crefname{section}{Section}{Sections}
\Crefname{table}{Table}{Tables}
\crefname{table}{Tab.}{Tabs.}

\makeatletter
\DeclareRobustCommand\onedot{\futurelet\@let@token\@onedot}
\def\@onedot{\ifx\@let@token.\else.\null\fi\xspace}

\def\eg{\emph{e.g}\onedot} 
\def\ie{\emph{i.e}\onedot}

\makeatother

\newcommand{\boldparagraph}[1]{\vspace{0pt}\noindent{\bf #1}}
\newcommand{\ours}{3D-AVS\xspace}  %
\newcommand{\scannet}{ScanNet\xspace}
\newcommand{\nuscenes}{nuScenes\xspace}

\definecolor{turquoise}{cmyk}{0.65,0,0.1,0.3}
\definecolor{purple}{rgb}{0.65,0,0.65}
\definecolor{dark_green}{rgb}{0, 0.5, 0}
\definecolor{orange}{rgb}{0.8, 0.6, 0.2}
\definecolor{red}{rgb}{0.8, 0.2, 0.2}
\definecolor{darkred}{rgb}{0.6, 0.1, 0.05}
\definecolor{blueish}{rgb}{0.0, 0.3, .6}
\definecolor{light_gray}{rgb}{0.7, 0.7, .7}
\definecolor{pink}{rgb}{1, 0, 1}
\definecolor{greyblue}{rgb}{0.25, 0.25, 1}

\colorlet{colorFst}{Green!25}       %
\colorlet{colorSnd}{SpringGreen!45} %
\colorlet{colorTrd}{Yellow!30}      %
\colorlet{colorLow}{darkgray!30}    %
\newcommand{\fs}{\cellcolor{colorFst}\bf}   %
\newcommand{\nd}{\cellcolor{colorSnd}}      %
\newcommand{\rd}{\cellcolor{colorTrd}}      %

 \iftrue  %
\newcommand{\fkn}[1]{[{\bf \color{purple} Fatemeh: #1}]}
\newcommand{\Weijie}[1]{[{\bf \color{blue} Weijie: #1}]}
\newcommand{\ozzy}[1]{[{\bf \color{ForestGreen} Ozzy: #1}]}
\newcommand{\mo}[1]{[{\bf \color{Orange} Martin: #1}]}
\newcommand{\TODO}[1]{\textbf{\color{red}[TODO: #1]}}

\else 
\newcommand{\fkn}[1]{}
\newcommand{\Weijie}[1]{}
\newcommand{\ozzy}[1]{}
\newcommand{\mo}[1]{}
\newcommand{\TODO}[1]{}

\fi
\newcommand{\xmark}{\ding{55}}%
\newcommand{\noo}{\textcolor{red}{\xmark}}
\newcommand{\yes}{\textcolor{OliveGreen}{\checkmark}}

\makenomenclature
\renewcommand\nomgroup[1]{%
  \item[\bfseries
  \ifstrequal{#1}{M}{Method variables}{%
  \ifstrequal{#1}{E}{Evaluation}{%
  \ifstrequal{#1}{S}{Scalars}{}}}%
]}

\newcommand{\nomenclcustom}[3]{\nomenclature[#1#2]{#3}}

\newcommand{\beginsupplement}{%
        \setcounter{table}{0}
        \renewcommand{\thetable}{S\arabic{table}}%
        \setcounter{figure}{0}
        \renewcommand{\thefigure}{S\arabic{figure}}%
        \setcounter{section}{0}
        \renewcommand{\thesection}{S\arabic{section}}%
    }

%% file: sec/0_abstract.tex
\begin{abstract}
Open-Vocabulary Segmentation (OVS) methods offer promising capabilities in detecting unseen object categories, but the category must be known and needs to be provided by a human, either via a text prompt or pre-labeled datasets, thus limiting their scalability. We propose \ours, a method for Auto-Vocabulary Segmentation of 3D point clouds for which the vocabulary is unknown and auto-generated for each input at runtime, thus eliminating the human in the loop and typically providing a substantially larger vocabulary for richer annotations. \ours first recognizes semantic entities from image or point cloud data and then segments all points with the automatically generated vocabulary. Our method incorporates both image-based and point-based recognition, enhancing robustness under challenging lighting conditions where geometric information from LiDAR is especially valuable.
Our point-based recognition features a Sparse Masked Attention Pooling (SMAP) module to enrich the diversity of recognized objects.
To address the challenges of evaluating unknown vocabularies and avoid annotation biases from label synonyms, hierarchies, or semantic overlaps, we introduce the annotation-free Text-Point Semantic Similarity (TPSS) metric for assessing generated vocabulary quality. Our evaluations on nuScenes and ScanNet200 demonstrate \ours's ability to generate semantic classes with accurate point-wise segmentations.

\end{abstract}

%% file: sec/1_intro.tex
\section{Introduction}
\label{sec:intro}

Existing perception methods~\cite{aTao_2020_Hierarchical, sBorse_2021_InverseForm, zLi_2022_BEVFormer, yYan_2018_SECOND, tYin_2021_CenterPoint, lang2019pointpillars} for autonomous driving often rely on an inclusiveness assumption that all potential categories of interest must exist in the training dataset.
Nevertheless, public datasets often annotate instances with pre-defined categories, which can vary from three (\eg vehicle, cyclist and pedestrian)~\cite{pSun_2020_Waymo, jMao_2021_ONCE} to several dozen types~\cite{nuscenes, SemanticKITTI}, and fail to annotate rare objects with correct semantic labels.
Failing to recognize atypical objects or road users poses a significant risk to the perception model's adaptability to diverse real-life scenarios.

\input{figures/fig1}

The development of Vision-Language Models (VLMs) strengthens the connection between vision and language modalities and promotes progress in multi-modal tasks, such as
zero-shot image classification~\cite{aRadford_2021_CLIP}, image search and retrieval~\cite{aRamesh_2021_DALLE}, image captioning~\cite{clipcap}, video understanding~\cite{hXu_2021_VideoCLIP}, and open-vocabulary learning~\cite{jWu_2024_OV_Survey_PAMI}.
Open-vocabulary learning methods often utilize pre-trained VLMs to find the correspondence between visual entities and a semantic vocabulary, thereby creating the potential to detect any category of interest~\cite{jWu_2024_OV_Survey_PAMI, xZhou_2023_VisionLanguageModels}.
However, these methods rely on human-specified queries, and thus can not dynamically recognize all semantic entities in a scene. Conversely, predefining everything is neither scalable nor practical in a dynamic world, as it is impossible to anticipate all the categories the model may encounter in advance.
This shortcoming severely limits the real-life applicability of existing methods, as newly encountered object categories could still be unknown to the model or unaware to humans.

In this work, we propose \ours, a framework that automatically recognizes objects, generates a vocabulary for them and segments LiDAR points.
We evaluate our method on indoor and outdoor datasets~\cite{nuscenes, dai2017scannet, scannet200} and introduce a metric, TPSS, to assess the model performance based on semantic consistency in CLIP~\cite{aRadford_2021_CLIP} space.
\Cref{fig:human_vs_ours} compares the same segmenter, namely OpenScene~\cite{sPeng_2023_OpenScene} with different vocabularies.
\ours generates convincing semantic classes as well as accurate point-wise segmentations.
Moreover, when pre-defined categories are general and ambiguous, \eg \textit{man-made}, \ours recognizes the semantically more precise categories, \eg \textit{building} and \textit{signboard}.

Our contributions can be summarized as follows:
\textbf{1)} we introduce auto-vocabulary segmentation for point clouds, aiming to label all points using a rich and scene-specific vocabulary. Unlike methods that rely on predefined vocabularies, we address an \emph{unknown} vocabulary setting by dynamically generating vocabulary per input;
\textbf{2)} we propose \ours, a framework that automatically identifies objects, either through an image-free point-based captioner or an off-the-shelf image-based captioner; 
\textbf{3)} we propose a point captioner for \ours-LiDAR that decodes text from point-based CLIP features, achieving image independence and enhanced object diversity through a sparse masked attention pooling (SMAP) module; and
\textbf{4)} we introduce the Text-Point Semantic Similarity score, a novel CLIP-based, annotation-free metric that evaluates semantic consistency, accounting for synonyms, hierarchies, and similarity in unknown vocabularies, enabling scalable auto-vocabulary evaluation without human input.

%% file: figures/fig1.tex
\begin{figure}[t] \centering
    \includegraphics[width=0.46\textwidth]{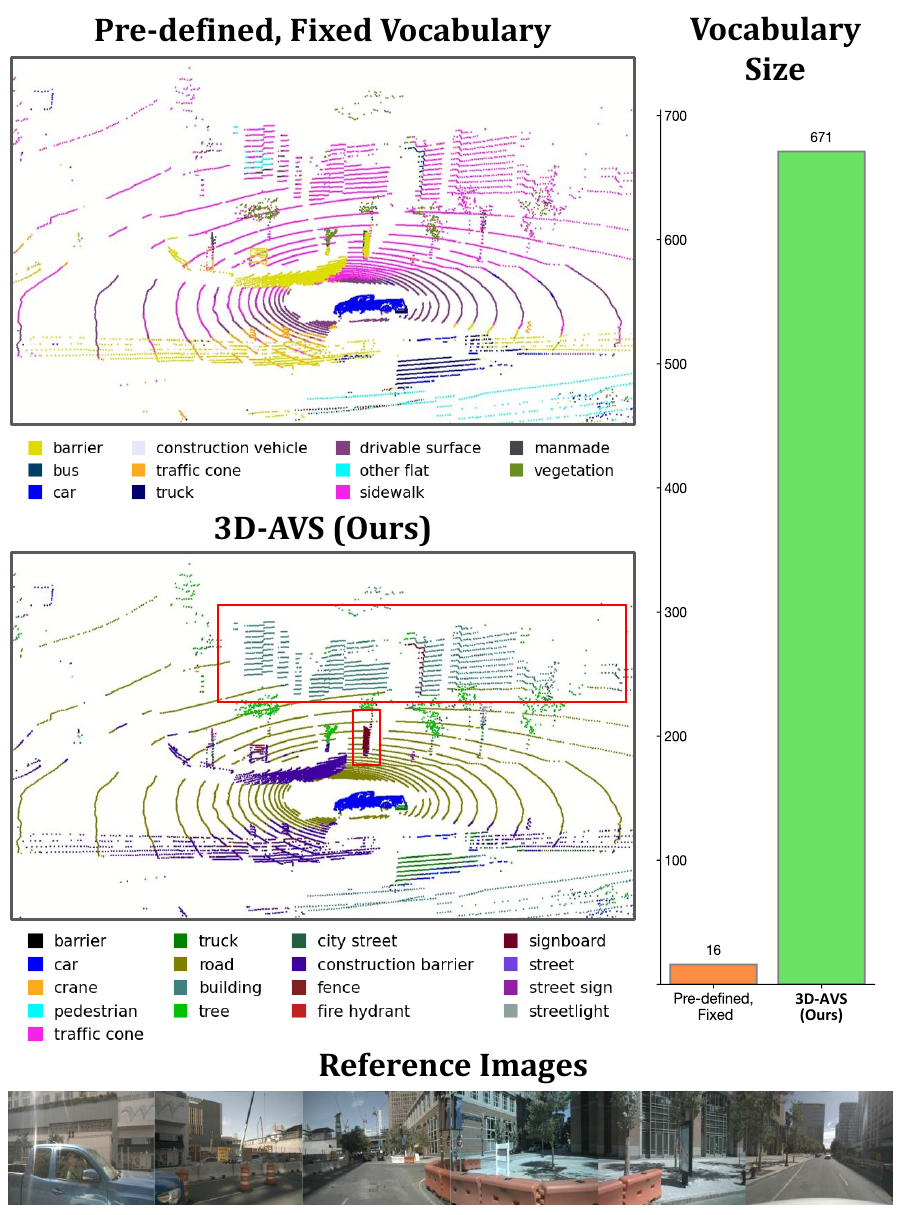}
    \caption{\textbf{Pre-defined Vocabulary vs. Auto-Vocabulary.} \ours automatically generates a vocabulary for which it predicts segmentation masks, offering greater semantic precision. Our predictions identify specific classes \eg \textit{building} and \textit{signboard} (highlighted in red boxes), which are annotated with ambiguous terms like \textit{manmade}. Quantitatively, \ours recognizes 671 unique categories on the validation set of \nuscenes~\cite{nuscenes}, significantly surpassing \nuscenes's original 16 categories. \textbf{Left:} Vocabulary for a single scene, \textbf{Right:} Vocabulary for the entire dataset.}
    \label{fig:human_vs_ours}
    \vspace{-1em}
\end{figure}

%% file: sec/2_related_work.tex
\section{Related Work}
\label{sec:related_works}

\boldparagraph{Open-Vocabulary Segmentation (OVS).} \label{sec:ovs}
OVS aims to perform segmentation based on a list of arbitrary text queries. 
CLIP~\cite{aRadford_2021_CLIP} achieves this in 2D by aligning vision and language in a shared latent space. However, no comparable large-scale point cloud dataset exists for similar training in 3D.
Additionally, captions in point cloud datasets are typically much sparser.
Therefore, existing methods usually freeze the text encoder and image encoder, and align point features to vision-language feature space~\cite{lXue_2023_ULIP, sPeng_2023_OpenScene, rChen_2023_CLIP2Scene, yZeng_2023_CLIP_2, mNajibi_2023_UP-VL}.
ULIP~\cite{lXue_2023_ULIP} distils vision-language knowledge into a point encoder via contrastive learning on text-image-point triplets.
CLIP2Scene~\cite{rChen_2023_CLIP2Scene} adopts self-supervised learning, aligning point-text features using spatial-temporal cues.
OpenScene~\cite{sPeng_2023_OpenScene} supervises the point encoder with CLIP-based image features through point-pixel projection. 
While these OVS approaches show promising results, they require user-defined categories as prompts. 
Conversely, our approach automatically generates categories that potentially appear in the scene without any human in the loop.

\boldparagraph{Auto-Vocabulary Segmentation (AVS).} \label{sec:avs}
AVS differs from OVS in that it segments entities directly from perceptual data rather than relying on a human-defined vocabulary as input.
Relevant target categories are directly inferred from the image - usually without any additional training, finetuning, data sourcing or annotation effort.
Zero-Guidance Segmentation (ZeroSeg)~\cite{pRewatbowornwong_2023_ZeroSeg} achieve this by using clustered DINO~\cite{mCaron_2021_DINO} embeddings to obtain binary object masks.
These masks were used to guide the attention of CLIP, resulting in embeddings that are more accurately targeted to individual segments, and a trained large language model was tasked to output texts closest to said embeddings.
While this required switching between three different latent representations, AutoSeg~\cite{oUlger_2023_AutoSeg} proposed a more direct approach based on BLIP~\cite{jLi_2022_BLIP} embeddings only.
They introduced a procedure in which multi-scale BLIP embeddings are enhanced through clustering, alignment and denoising.
The embeddings are then captioned using BLIP's decoder and parsed into a noun set used by an OVS model for segmentation.
CaSED~\cite{aConti_2024_CaSED} retrieves captions from an external database and then integrates parsed texts with different segmentation methods.
Despite these attempts in 2D domain, AVS in 3D domain remains unexplored.
Concurrently and independently, Meng et al.~\cite{gMei_2025_PoVo} have proposed vocabulary-free 3D instance segmentation and a method PoVo for this task.
While PoVo first obtains 3D clusters and then matches the generated semantic categories to the clusters, our work focuses more on target category generation and seamless integration with existing OVS methods.

\boldparagraph{Challenges of AVS Evaluation.} \label{sec:sec2_avs_challenges}
AVS presents additional challenges linked to evaluation.
Since generated categories can be open-ended and outside of the fixed dataset vocabulary, one needs to bridge the gap between the two to assess the segmentation performance.
ZeroSeg~\cite{pRewatbowornwong_2023_ZeroSeg} exploits subjective assessment.
In AutoSeg~\cite{oUlger_2023_AutoSeg}, the LLM-based mapper, LAVE, is introduced to address this challenge. However, the mapping targets are typically limited in size, causing the auto-generated categories - often more semantically rich and precise - to be discarded.
To overcome these limitations, we propose the TPSS metric, which enables the evaluation of the generated categories while preserving their open-ended nature.

\boldparagraph{Captioning 2D and 3D Data.} \label{sec:relatedwork_captioning}
Captioning is the process of generating a concise and meaningful description from data modalities such as images, videos or point clouds. Notable works in 2D combine image-based templates with extracted attributes~\cite{babytalk, imageparsing}, combine deep learning models like convolutional neural networks with RNN, LSTM or transformer-based generators~\cite{showandtell, showandtellicml, you2016imagecaptioningsemanticattention}, or leverage pre-trained vision-language embeddings such as CLIP or BLIP~\cite{clipcap, clips, llava, jLi_2022_BLIP}.
BLIP~\cite{jLi_2022_BLIP}, known for its effective but somewhat generic captions, often focus only on the 2-3 most prominent entities in an image.
BBoost~\cite{oUlger_2023_AutoSeg} addresses this limitation by enhancing BLIP tokens through unsupervised semantic clustering in the latent space, enabling cluster-wise captioning and resulting in more comprehensive and detailed captions.
More recently, xGen-MM (BLIP-3)~\cite{blip3} was introduced, building on BLIP with two improvements: an expanded and more diverse set of training data, and a scalable vision token sampler for flexible input resolutions.
While this task is broadly explored in the 2D domain, it is yet to be solved in the 3D domain.
Existing approaches focus on describing a single object, \eg CAD models~\cite{lXue_2023_ULIP, lXue_2023_ULIP-2} and scanned shapes~\cite{tLuo_2023_Cap3D, lXue_2023_ULIP-2, zGuo_2023_Point-Bind, rXu_2024_PointLLM, zZhu_2023_3D-VisTA}, or dense contextual indoor scenarios~\cite{yHong_2023_3D-LLM, dzChen_2023_UniT3D, yChen_2024_Grounded3DLLMReferent, zWang_2023_Chat-3D, hHuang_2023_Chat-3DV2, sChen_2024_LL3DA, zZhu_2023_3D-VisTA}, but fail to caption sparse outdoor scenes due to sparsity and lack of colour information.
LidarCLIP~\cite{gHess_2024_LidarCLIP} encodes a sparse point cloud to a CLIP feature vector and then decodes it to a caption via ClipCap~\cite{rMokady_2021_ClipCap}.
However, LidarCLIP only provides a global caption per scene, leading to limited coverage of semantic entities.
Instead, our proposed point captioner copes with flexible receptive fields and offers a controllable number of captions with various granularity.

%% file: sec/3_methodology.tex
\input{figures/overview_autovoc}
\section{Method} \label{sec:method}

\subsection{Preliminaries} \label{sec:preliminaries}
\boldparagraph{CLIP and CLIP-aligned Encoder.}
CLIP~\cite{aRadford_2021_CLIP} is believed to properly align visual and text features due to its superior performance on vision-language tasks.
It comprises a text encoder $h_\text{tx}$ and an image encoder $h_\text{im}$, both of which map a data modality, \eg text and image, to a vision-language latent space,  also known as the CLIP space.
Many works~\cite{cho_cvpr2024_catset, xu_cvpr23_SAN, yu2023fcclip, LSeg, openseg} increase the output resolution of CLIP image encoder, yielding high-resolution features $h_\text{im}^\text{hr}$, while preserving alignment within the original CLIP space.
Furthermore, some 3D methods~\cite{lXue_2023_ULIP, sPeng_2023_OpenScene, mNajibi_2023_UP-VL, gHess_2024_LidarCLIP} distill features from $h_\text{im}$ or its high-resolution variant $h_\text{im}^\text{hr}$ into 3D backbones, yielding CLIP-aligned 3D encoder $h_\text{pt}$.
In this paper, we leverage such aligned 3D encoders and bypass the time-consuming training process whenever possible.

\boldparagraph{Problem Definition.} \label{sec:problem_definition}
Given a point cloud $\mathbf{P}=\{p_n\}_{n=1}^N \in \mathbb{R}^{N \times 3}$ with $N$ points, the aim is to assign a semantic class label $l \in \mathbb{S}$ to every point, where $\mathbb{S}$ indicates a vast semantic space.
Different to closed-set or open-vocabulary segmentation for which the vocabulary is \emph{known} either via a user-specified prompt or by pre-defined labels from dataset, the class set in auto-vocabulary segmentation is \emph{unknown} and automatically generated for each input scene.

\subsection{3D Auto-Vocabulary Segmentation} \label{sec:autovoc3d}
This section introduces \ours for which an overview of its major components is shown in \cref{fig:autovoc_overview}. 
Given a point cloud and a set of corresponding images, \ours first utilizes a point captioner and an image captioner to describe points and images in detail.
The generated captions are parsed in the Caption2Tag module, resulting in a list of tags indicating semantic entities.
Eventually, each point is assigned a semantic tag, forming segmentation results.
These key components are elaborated in the following paragraphs.

\boldparagraph{Scene Captioning.}
A key step of our approach is the auto-generation of a vocabulary for the given scene, which is performed by a scene captioner that is either based on input images or on the input point cloud.
Image captioning is a well-explored task with a variety of accessible multi-modality large-language models (MLLMs)~\cite{blip3, jLi_2022_BLIP, zhang2023ram}.
We adopt xGen-MM~\cite{blip3} as the image captioner because of its architectural flexibility and enhanced semantic coverage, 
Given a set of \( K \) images \( \mathbf{I} \in \mathbb{R}^{K \times H \times W \times 3} \) capturing a scene, and an instruction prompt (details in supplementary material), the image captioner generates a list of captions
\begin{equation}
    \mathbf{D} = \left\{ \dvect_{\text{im}}^{(k)} \in \mathbb{R}^{w_k} \mid k = 1, \dots, K \right\}
\end{equation}
where $w_k$ is the number of words in the caption for the $k$-th image. To ensure a diverse enough set of coherent captions, we opt for beam search in the generation process. Implementation details are in the supplementary material. Following caption generation, each caption is parsed and validated with Caption2Tag, as described in the section below.

LiDAR point cloud captioning remains an underexplored area in existing research despite the potential of such captions for applications.
While images collected alongside LiDAR point clouds can be used to generate a target vocabulary, relying solely on images proves inadequate under challenging conditions such as low light or adverse weather, where visual data becomes unreliable. To address this, we introduce a novel Point Captioner trained via transfer learning, which provides captions directly from color-independent LiDAR data. Our approach, detailed in \cref{sec:point_captioner}, takes a point cloud \( \mathbf{P} \) as input and outputs captions \( \dvect_{\text{pt}} \). 
Unlike image captioning, which requires extensive contextual information and sophisticated vision models to produce detailed captions, the Point Captioner provides robust descriptions by relying solely on geometric features. This color independence is particularly beneficial in low-visibility environments, such as nighttime scenes where image-based captioning often falls short. Combining both modalities ultimately yields the best results, uniting the diversity of image captions with the resilience of point-based captions.

\boldparagraph{Text Parsing.} Captions generated by the image and point captioner are scene-specific sentences in natural language which we then parse into individual object nouns for semantic segmentation. To this end, we filter the sentence on (compound) nouns (\ie general entities) and proper nouns (\ie named entities) using spaCy~\cite{spacy} and transform them to their singular form through lemmatization. Lastly, we verify each category against the WordNet dictionary, resulting in a set of $M$ scene-specific tags, denoted as $\mathbf{L}=\{l_m\}_{m=1}^M$. %

\boldparagraph{Segmentation.}
The proposed pipeline separates the vocabulary generation and segmentation, enabling seamless integration with an open-vocabulary point segmenter.
The segmenter consists of three encoders, namely a text encoder $h_\text{tx}: \mathbb{R}^1 \rightarrow \mathbb{R}^C$, a high-resolution image encoder $h^\text{hr}_\text{im}: \mathbb{R}^{H \times W \times 3} \rightarrow \mathbb{R}^{H \times W \times C}$ and a point encoder $h_\text{pt}: \mathbb{R}^{N \times 3} \rightarrow \mathbb{R}^{N \times C}$, that are pre-aligned with the CLIP vision-language latent space.
Following the inference procedure of CLIP~\cite{aRadford_2021_CLIP}, namely similarity-based label assignment, we first compute the embeddings as follows:
\begin{align}			    
  \mathbf{E}_\text{tx} &=\{e_m\}_{m=1}^M \leftarrow h_\text{tx}(\mathbf{L}) \\
  \mathbf{F}_\text{im} &=\{f_k\}_{k=1}^K \leftarrow h_\text{im}(\mathbf{I}) \\
  \mathbf{F}_\text{pt} &=\{f_n\}_{n=1}^N \leftarrow h_\text{pt}(\mathbf{P})
  \label{eq:encode}
\end{align}
where $\mathbf{E}_\text{tx}$, $\mathbf{F}_\text{im}$ and $\mathbf{F}_\text{pt}$ indicate text embeddings, image features, and point features.
$e_m \in \mathbb{R}^{1 \times C}$, $f_k \in \mathbb{R}^{H \times W \times C}$ and $f_n \in \mathbb{R}^{1 \times C}$ represent per-label, per-image and per-point features. Then, the image features are lifted to 3D and assign each point a pixel feature if the point is visible in the images.
In other words, given a point, we calculate its 2D coordinates by point-to-pixel mapping $\varGamma: \mathbb{R}^3 \rightarrow \mathbb{R}^2$ and then copy the corresponding pixel feature to the point, denoted as $f_n^\text{im} \in \mathbb{R}^{1 \times C}$. %
Eventually, each point is assigned a semantic label as follows:
\begin{align}
    \hat{l}_n=\underset{m}{\mathrm{argmax}} \left ( \mathrm{max} \left ( \mathrm{SIM} (f_n, e_m) || \mathrm{SIM} (f_n^\text{im}, e_m \right ) \right )  
\end{align}
where $\hat{l}_n$ denotes the predicted label for point $p_n$, $\mathrm{SIM}(\cdot,\cdot)$ is a similarity metric, for which we employ dot product, producing a tensor $\in \mathbb{R}^{1 \times M}$ and $||$ indicates concatenation when image features are available.
$\mathrm{max}(\cdot)$ takes a tensor $\in \mathbb{R}^{2 \times M}$ as input, performs a column-wise maximum operation, and outputs a tensor $\in \mathbb{R}^{1 \times M}$.

\input{figures/point_captioner}
\subsection{Point Captioner} \label{sec:point_captioner}
Inspired by LidarClip~\cite{gHess_2024_LidarCLIP}, we develop the Point Captioner that first encodes points to CLIP latent space and then decodes CLIP features to captions.
However, LidarClip only provides a global caption per point cloud, leading to limited coverage of semantic entities.
Therefore, we propose a sparse masked attention pooling (SMAP) that can increase the receptive field and output a controllable number of feature vectors, making it possible to train the network with a varying number of images.
We detail the training stage, the inference stage and the SMAP in the following paragraphs.

\boldparagraph{Training.}
The training of the Point Captioner is essentially a 2D-to-3D distillation that transfers knowledge from the 2D vision foundation model to the 3D backbone.
We utilize the CLIP image encoder~\cite{aRadford_2021_CLIP} $h_\text{im}^\text{clip}: \mathbb{R}^{H \times W \times 3} \rightarrow \mathbb{R}^{1 \times 1 \times C}$ and a CLIP-aligned point encoder $h_\text{pt}: \mathbb{R}^{N \times 3} \rightarrow \mathbb{R}^{N \times C}$ to encode images and points.
However, $h_\text{im}^\text{clip}(\cdot)$ outputs a global feature vector that does not match the per-point features obtained from $h_\text{pt}(\cdot)$.
Therefore, we add SMAP to pool point-wise features.
As shown in \cref{fig:point_captioner} (left), during training, a point cloud and a point-to-pixel mapping function (visualized as an image) are fed to the image-based mask generation.
The output is a point-wise binary mask, where \textit{true} indicates the point is visible in the image.
We visualize the point mask by projecting the point to the image.
The mask and the point features obtained from $h_\text{pt}(\cdot)$ are input to SMAP.
SMAP integrates features of points that are visible in the image and is supervised by the output feature of $h_\text{im}^\text{clip}(\cdot)$.
Note that only one image is visualized in \cref{fig:point_captioner} for clarity but all images corresponding to the point cloud are processed in parallel during training.

\boldparagraph{Inference.}
Our goal is to generate diverse captions that comprehensively cover all semantic categories without requiring the intrinsic parameters of cameras.
To achieve this, we propose a \textbf{geometry-based mask generation} strategy that efficiently partitions the point cloud into multiple regions, followed by individual captions for each region.
Given the differences in point cloud distributions, we adopt cylindrical sector-based partitioning for outdoor scenes and square pillar-based partitioning for indoor scenes.
In the remainder of this paragraph, we illustrate our approach using outdoor point clouds as an example, while details on indoor partitioning are provided in the supplementary materials.
The point cloud is first transformed from a Cartesian coordinate system $\{p_n = (x_n, y_n, z_n)\}_{n=1}^N$ to a polar coordinate system $\{p_n = (\rho_n, \varphi_n, z_n)\}_{n=1}^N$ and then split into $T$ sectors according to its polar angle $\varphi$.
The binary masks $\mathcal{B}=\{b_n^t\} \in \mathbb{R}^{N \times T}$ are obtained as follows:
\begin{align}
    b_n^t = 
    \begin{cases}
        \text{true},  & \text{if } \frac{t}{T} 2 \pi \leq \varphi < \frac{t+1}{T} 2\pi \\
        \text{false}, & \text{otherwise.}
    \end{cases}
\end{align}
where $t \in \{0, 1, \ldots, T-1\}$.
This way, SMAP generates mask-wise features that are further decoded into captions in the caption module.
The merit of this method is that the number of captions is controllable by changing $T$.

\input{figures/sparse_masked_atten_pooling}

\boldparagraph{Sparse Masked Attention Pooling (SMAP).}
SMAP takes as input 1) an entire point cloud with its per-point coordinates $\mathcal{C} \in \mathbb{R}^{N \times 3}$ and features $\mathcal{F} = \mathbf{F}_\text{pt} \in \mathbb{R}^{N \times C}$, and 2) $J$ binary point-wise masks $\mathcal{B} \in \mathbb{R}^{J \times N}$, where $J=K$ during training and $J=T$ during inference.
SMAP first conducts a relative positional encoding and then applies the masks to the encoded point features:
\begin{align}
    \mathcal{F}' &= \mathcal{B} * \big( \mathcal{F} + \mathrm{PE}(\mathcal{C}, \mathcal{F}) \big) 
\end{align}
where $\mathrm{PE}$ indicates a relative positional encoding as in~\cite{xWu_2024_PTv3} and $*$ denotes matrix multiplication. %
The masks essentially divide a point cloud into several subsets, allowing replacement.
Therefore, the feature $\mathcal{F}' = \{f'_j\}_{j=1}^J \in \mathbb{R}^{N_j \times C}$ has a variable length per mask.
After multiplication, features $F'$ go through two paths: 
1) zero-padded to the same length and then delivered to multi-head attention (MHA) as key $\mathcal{K}$ and value $\mathcal{V}$, and 2) passed to a global average pooling and then input to MHA as query $\mathcal{Q}$.
Eventually, we obtain pooled features $\mathcal{F}'' \in \mathbb{R}^{J \times C}$.

%% file: figures/overview_autovoc.tex
\begin{figure*}[t] 
\centering
    \includegraphics[width=0.98\textwidth]{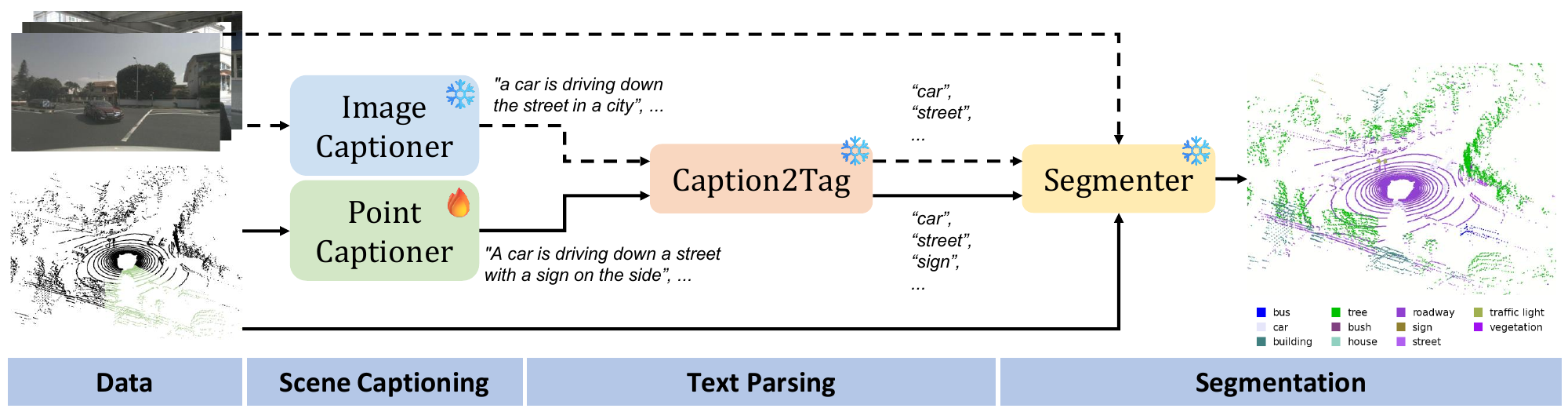}
    \caption{\textbf{Overview of \ours.} A point cloud and corresponding images are fed to respective point captioner and image captioner to generate captions. Then, Caption2Tag excludes irrelevant words in the captions. The remaining nouns are passed to a text encoder and eventually assigned to points through a segmenter. The dashed lines indicate that the entire images branch is optional. The point captioner is the only trainable component in \ours. Note that, the example point caption is generated based on observing the green points.
    }
    \label{fig:autovoc_overview}
    \vspace{-1em}
\end{figure*}

%% file: figures/point_captioner.tex
\begin{figure*}[t] \centering
    \includegraphics[width=0.95\textwidth]{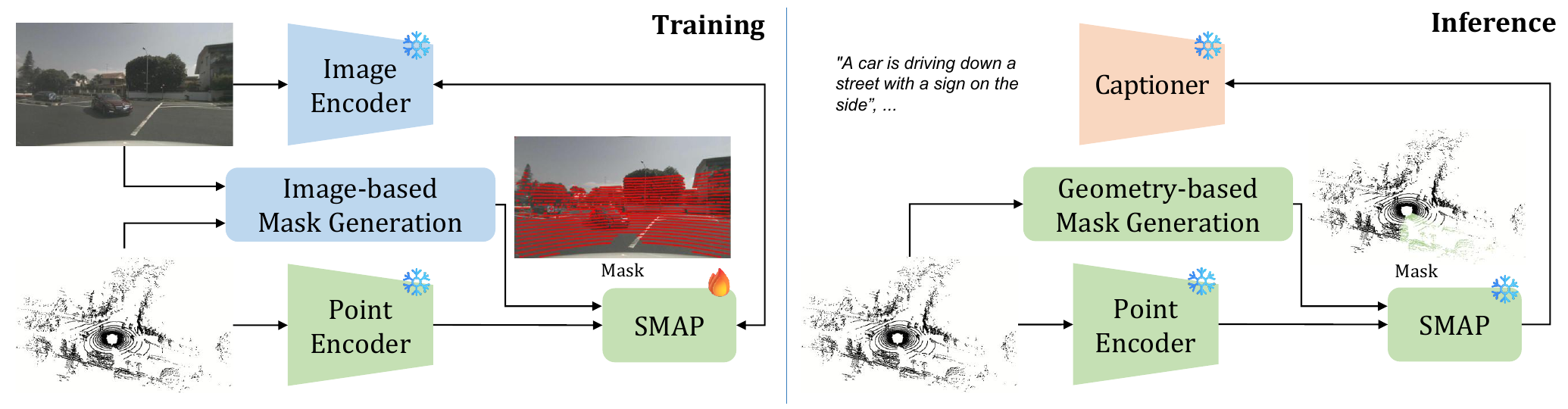}
    \caption{\textbf{Point Captioner Overview.} The image encoder and point encoder are pre-aligned in the CLIP latent space. During training \textbf{(left)}, Sparse Masked Attention Pooling (SMAP) aggregates features from points visible in the image (highlighted in red) and is supervised using CLIP image features. During inference \textbf{(right)}, neither the image nor camera intrinsic parameters are available. To address this, a group of masks are generated based solely on geometric information. The SMAP output is then decoded into a group of captions. For simplicity, only one image (left) and one sector (right) are shown.}
    \label{fig:point_captioner}
    \vspace{-0.5em}
\end{figure*}

%% file: figures/sparse_masked_atten_pooling.tex
\begin{figure}[t] \centering
    \includegraphics[width=0.46\textwidth]{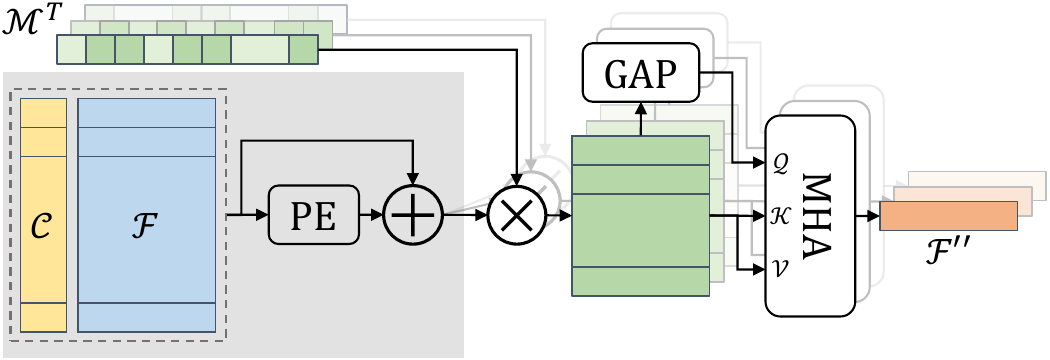}
    \caption{\textbf{Sparse Masked Attention Pooling (SMAP).} Given the coordinates and features of all points, a relative positional encoding (PE) is applied, followed by a residual connection. Masks are applied to the points, creating groups of point subsets. Global Average Pooling (GAP) on each subset produces a mean feature as a query. Finally, multi-head attention (MHA) is applied within each group to generate one feature per subset.}
    \label{fig:smap}
    \vspace{-0.8em}
\end{figure}

%% file: sec/4_evaluation.tex
\section{Evaluation} \label{sec:evaluation}
Auto-vocabulary segmentation introduces a novel setting without a standardized benchmark, making it challenging to compare methods directly.
In this section, we introduce the challenges of evaluation this novel task and propose two strategies to evaluate segmentation accuracy and the semantic consistency between points and text labels.

\subsection{Challenges} \label{sec:eval_challenges}
In open-vocabulary segmentation, which is a similar but simpler task, evaluation can be performed on conventional segmentation datasets by using the categories present in the annotations as pre-defined queries. However, this inherently means the model has prior knowledge of the classes it is expected to predict. In auto-vocabulary segmentation, however, no such information is available beforehand, presenting a unique challenge for evaluation. Moreover, natural language introduces ambiguities~\cite{oUlger_2023_AutoSeg, YADAV202185}, creating complex relationships between classes, such as synonymy, hyponymy and hypernymy. For instance, \textit{road} could be labeled as \textit{drivable surface}, \textit{street}, or \textit{roadway}, while a \textit{tire} might be classified independently or as part of a wheel or vehicle. This makes it challenging to determine whether an instance is appropriately tagged with a precise semantic label. Given these nuances, evaluating the quality of generated labels and segmentation accuracy becomes complex, as the model must align with the varying language used in annotations, even when sometimes only general categories are provided in the ground truth.

To address these challenges, we propose two solutions. Firstly, we introduce a novel, objective and annotation-independent metric in \cref{sec:TPSS} that assesses how accurately a label - either auto-generated or selected from a fixed vocabulary - fits a given 3D point. This metric allows for flexible, any-to-any class evaluation. Secondly, we leverage an LLM-based mapping approach to align auto-generated vocabulary classes with the ground-truth classes, enabling us to effectively evaluate both the quality of the segmentation mask and the relevance of the predicted labels (\cref{sec:lave}).

\subsection{Text-Point Semantic Similarity Metric} \label{sec:TPSS} 
We introduce the Text-Point Semantic Similarity (TPSS) metric, a measure independent of dataset annotations and subjective assessment.
TPSS draws inspiration from inference with CLIP~\cite{aRadford_2021_CLIP}, where the best label out of a set of target classes $\{m_0, ..., m_M\}$ is assigned to an image:
\begin{align}
  \hat{l} &= \underset{m}{\mathrm{argmax}} \big( \mathrm{SIM} (f^\text{im}, e_m) \big)
  \label{eq:clip_infer}
\end{align}
where $\hat{l}$ represents the predicted label, $f^\text{im}$ is the image feature, and $e_m$ denotes the text embeddings for class $m$. 
This equation identifies the label with the closest text embedding to the provided image feature in latent space, indicating the highest semantic alignment within CLIP's language space. TPSS metric employs a similar approach, comparing pairs of individual point features with text features in this aligned space. This enables evaluation of how well any label corresponds to a specific point based on semantic similarity, making TPSS ideal for assessing both dynamic and fixed vocabularies.
For further illustration, consider a scenario where a LiDAR point belongs to an object outside the \nuscenes official classes, such as a ``trash bin'', and is thus annotated as ``background''. If our method predicts ``garbage can'' for this point, it should not be penalized for not predicting ``background'', as the original prediction is semantically closer to ``trash bin''. TPSS accounts for such cases, evaluating the predicted label based on the object's visual appearance rather than annotation setting or potential bias.
Formally, let $\mathbf{P}=\{p_n\}_{n=1}^N$ be a point cloud with $N$ points and $\mathbf{L}=\{l_m\}_{m=1}^M$ be a set of $M$ unique semantic labels generated for this point cloud.
The text embeddings $\mathbf{E}$ and the point features $\mathbf{F}$ are obtained as follows:
\begin{align}			    
  \mathbf{E}    &=  \{e_m\}_{m=1}^M \leftarrow g_\text{tx}(\mathbf{L}) \\
  \mathbf{F}    &=  \{f_n\}_{n=1}^N \leftarrow g_\text{pt}(\mathbf{P})
  \label{eq:tpss_encoder}
\end{align}
where $g_\text{tx}(\cdot)$ and $g_\text{pt}(\cdot)$ are the frozen CLIP text encoder~\cite{aRadford_2021_CLIP} and a CLIP-aligned point encoder, respectively.
The TPSS score is calculated as follows:
\begin{align}			    
  S_n &= \max_m \big( \mathrm{SIM}(f_n, e_m) \big) \\
  \mathrm{TPSS} (\mathbf{P}, \mathbf{L}, g_\text{tx}, g_\text{pt}) &= \underset{n}{\mathrm{mean}} ( S_n )
  \label{eq:tpss}
\end{align}
where $S_n$ is a point-wise similarity score for the point $n$.
$\mathrm{TPSS} (\mathbf{P}, \mathbf{L}, g_\text{pt}, g_\text{tx})$ measures the text-point semantic similarity between the point cloud $\mathbf{P}$ and the label set $\mathbf{L}$.
TPSS is encoder-agnostic as long as $g_\text{pt}$ and $ g_\text{tx}$ are aligned. However, to reliably quantify which label set aligns better with a given point cloud,
the point encoder and text encoder must remain unchanged across comparisons.

\subsection{Mapping Auto-Vocabulary to Fixed Vocabulary} \label{sec:lave}
While TPSS effectively measures semantic similarity within the embedding space, evaluating the quality of the resulting segmentations is crucial for meaningful assessment. This requires establishing a correspondence between open-ended classes and the ground truth classes. To achieve this, we employ an evaluation scheme that leverages an LLM-based mapper, inspired by the LLM-based Auto-Vocabulary Evaluator (LAVE) \cite{oUlger_2023_AutoSeg}. LAVE maps each unique auto-vocabulary category to a fixed ground truth class in the dataset. After segmenting the LiDAR point cloud using auto-vocabulary categories, each classification is updated according to this mapping. For example, points labeled as \textit{sedan} are reclassified under the \textit{car} category. This mapping enables evaluation of segmentation quality using the widely accepted mean Intersection-over-Union (mIoU) metric based on fixed-vocabulary categories, facilitating comparison with prior methods. Our evaluation framework extends LAVE by integrating mappings with GPT-4o and SBERT~\cite{sbert}. While we provide detailed results of all methods in the supplementary material, GPT-4o is used throughout the main experiments due to its superior mapping accuracy compared to both SBERT and LAVE's Llama-2-7B.

%% file: sec/5_experiments.tex
\section{Experiments} \label{sec:exp}

\input{tables/tpss}

\input{figures/tpss_per_whether}
\subsection{Experimental Setup} \label{sec:implementation}
Our method is evaluated on \nuscenes~\cite{nuscenes}, \scannet~\cite{dai2017scannet} and ScanNet200~\cite{scannet200} datasets.
\nuscenes dataset~\cite{nuscenes} is a comprehensive real-world dataset for autonomous driving research, capturing diverse urban driving scenarios from Boston and Singapore.
To increase the spatial density, we aggregate LiDAR points over a 0.5-second interval, focusing on the dataset’s LiDAR segmentation benchmark with 16 manually annotated categories.
Given the homogeneity often found in autonomous driving scenarios, we also assess \ours on the \scannet ~\cite{dai2017scannet} and ScanNet200~\cite{scannet200}.
\scannet dataset is an indoor dataset with 20 annotated classes.
ScanNet200 updates the annotations of ScanNet with more and finer-grained categories, \ie 200 categories, while keeping the input point clouds unchanged.
Due to space constraints, we refer to implementation details in the supplementary material, such as details on image captioner, segmenter and SMAP.

\subsection{Label Set Comparison} \label{sec:tpss_comparisons}
We compare the generated label set with the fixed, human-defined vocabulary classes in \cref{tab:tpss}.
OpenScene~\cite{sPeng_2023_OpenScene} manually create a more fine-grained vocabulary of 43 categories for the \nuscenes~\cite{nuscenes} (originally 16 categories) dataset, boosting the TPSS performance on the dataset from 7.39 to 8.70. Although the performance gain is impressive, \Cref{tab:tpss} demonstrates that \ours-generated labels are more semantically consistent with point clouds than manually defined labels, as \ours outperforms the predefined categories on both \nuscenes~\cite{nuscenes} and \scannet~\cite{dai2017scannet} datasets.
Additionally, \Cref{tab:tpss} demonstrates that combining text generation from both camera and LiDAR inputs, as done in \ours, improves text-point semantic similarity. This advantage stems from \ours' ability to adapt to scenes where one modality struggles. For instance, the image captioner often faces challenges in night scenes due to limited color information, while the point captioner continues to accurately describe relevant objects. This is further reflected in \cref{fig:tpss_barplot}, which shows that the point captioner proves especially useful in visually challenging scenes where the Image Captioner falls short.

\input{figures/qua_res}

\subsection{Segmentation Comparison} \label{sec:miou_comparisons}

\input{tables/quantitative}
For quantitative comparison, we employ LAVE~\cite{oUlger_2023_AutoSeg} to map all generated novel categories back to predefined categories.
Next, we calculate segmentation metrics, namely mean IoU (mIoU), on the validation sets of \nuscenes~\cite{nuscenes}, \scannet~\cite{dai2017scannet}, and ScanNet200~\cite{scannet200}.
Note that \ours does not have any access to the predefined categories during testing, which makes the segmentation task much harder.

\boldparagraph{Outdoor Dataset.}
\Cref{tab:quantitative} shows \ours generates better segmentation results on \nuscenes, confirming the effectiveness of \ours' open-ended recognition capabilities.
The segmentation performance mainly benefits from automatically generated categories for the ambiguous \nuscenes categories, such as \textit{driveable surface}, \textit{terrain}, and \textit{man-made}, achieving mIoU of 68.2, 41.4, and 55.4, respectively—substantially outperforming OpenScene~\cite{sPeng_2023_OpenScene} (see details in supplementary material).
Such an increase is expected, as \ours is able to generate much more specific namings for these overly general categories which can easily introduce noise.
\Cref{fig:qualitative_results} highlights some of these generated categories, such as \textit{man-made} being correctly recognized as \textit{staircase}, \textit{building} and \textit{glass door}. 

\boldparagraph{Indoor Datasets.}
\Cref{tab:quantitative} shows that \ours achieves a lower mIoU on ScanNet~\cite{dai2017scannet} compared to using a fixed vocabulary. This is likely due to the extensive range and variety of objects, where the generated labels must be mapped to a small and coarse-grained set of 20 dataset categories.
The state-of-the-art (SOTA) performance on ScanNet200~\cite{scannet200} further supports this argument.
Notably, the predictions of \ours remain identical on ScanNet and ScanNet200, as the input data are the same; the only difference lies in the evaluation vocabulary—mapping to 20 coarse categories in ScanNet versus 200 fine-grained categories in ScanNet200.
This shift in evaluation granularity introduces a more challenging task while allowing for a more faithful and detailed assessment of segmentation performance.
\ours achieves state-of-the-art results on ScanNet200, underscoring its effectiveness in open-ended 3D segmentation tasks.

\subsection{Ablation Study} \label{sec:ablations}
Ablation studies are conducted on the image captioner, point captioner and LAVE mapping to validate our design choices and hyperparameters.
The corresponding results are provided in the supplementary material.

%% file: tables/tpss.tex
\begin{table}[t]
\centering
\caption{\textbf{TPSS on the validation sets of nuScenes~\cite{nuscenes} and ScanNet~\cite{dai2017scannet}.} Two datasets are created with 16 and 20 official categories, respectively. OpenScene~\cite{sPeng_2023_OpenScene} extends the \nuscenes label set by manually defining 43 sub-categories. \ours outperforms these human-defined categories on both datasets, demonstrating its ability to generate a semantically more precise label set.}
\scalebox{0.85}{\tablestyle{4pt}{1.0}
\begin{tabular}{lccc}
\toprule
    \textbf{Label Set}                              & \textbf{Human-involved}    & \textbf{\nuscenes}~\cite{nuscenes} & \textbf{\scannet}~\cite{dai2017scannet} \\
\midrule
    Official label set                              & \yes              & 7.39 & 3.44 \\
    Extended label set~\cite{sPeng_2023_OpenScene}  & \yes              & 8.70 & -    \\
\midrule
    3D-AVS-Image                                    & \noo              & \rd 8.78 & \rd 3.49 \\
    3D-AVS-LiDAR                                    & \noo              & \nd 8.80 & \nd 3.71 \\
    3D-AVS                                          & \noo              & \fs 9.65 & \fs 3.78 \\
    \bottomrule
\end{tabular}
}
\label{tab:tpss}
\vspace{-1em}
\end{table}

%% file: figures/tpss_per_whether.tex
\begin{figure}[t] \centering
    \includegraphics[width=0.46\textwidth]{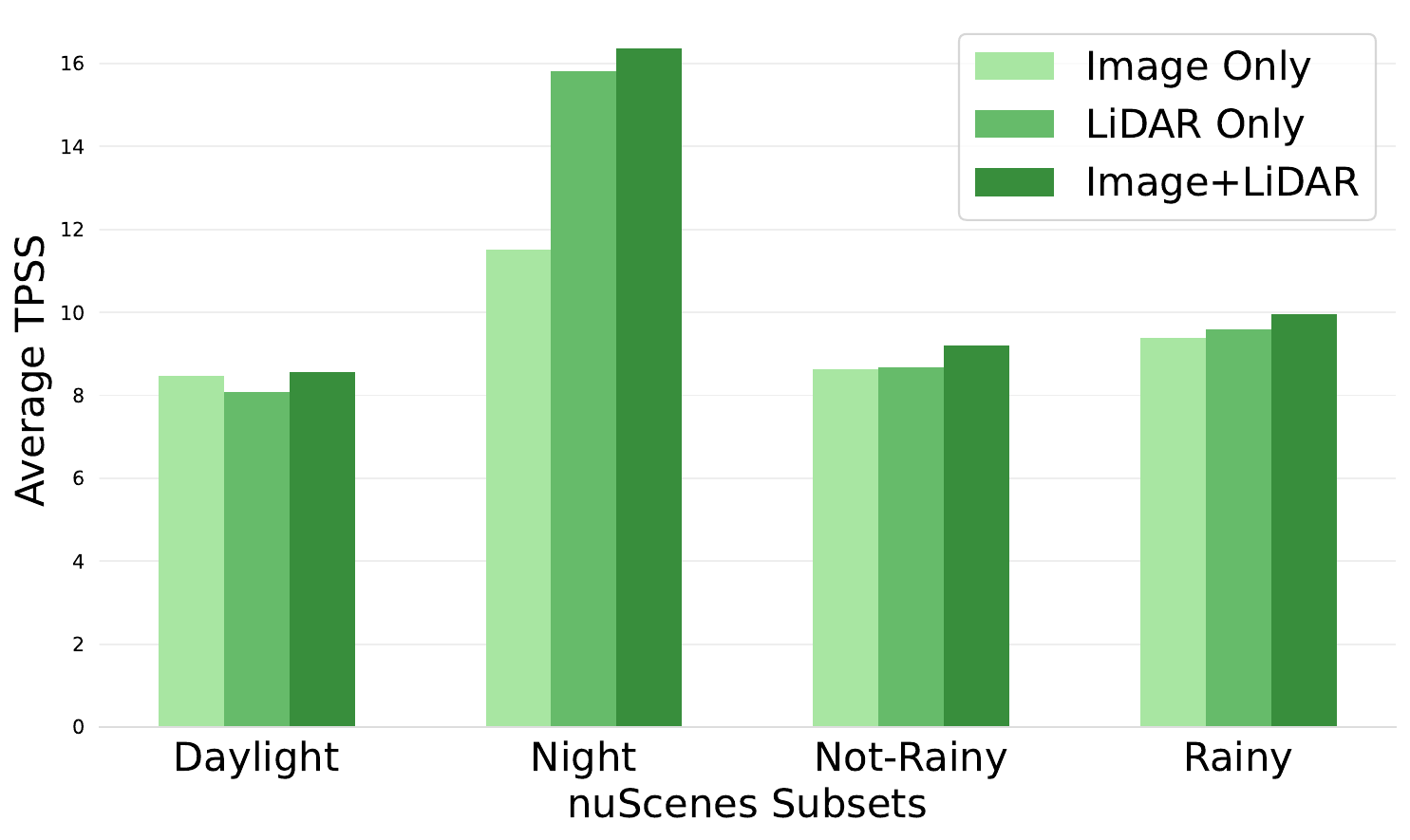}
    \caption{\textbf{TPSS on nuScenes subsets with different light conditions.} LiDAR-only \ours performs better during night and rainy scenes, suggesting its robustness across difficult conditions.}
    \label{fig:tpss_barplot}
    \vspace{-1em}
\end{figure}

%% file: figures/qua_res.tex
\begin{figure*}[t] 
\vspace{-10pt}
\centering
    \includegraphics[width=0.98\textwidth]{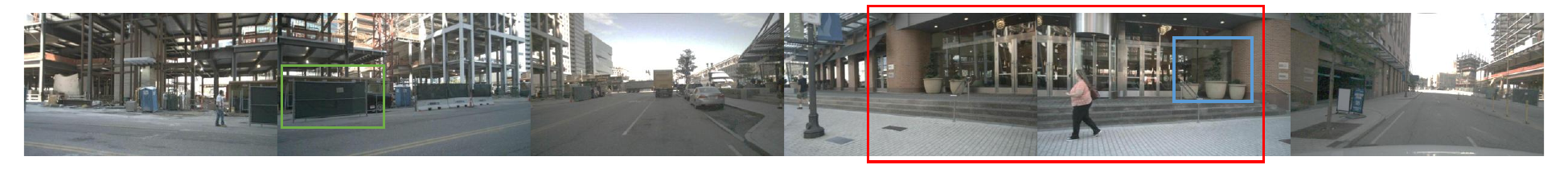}
    \\
    \vspace{-0.5em}
    \makebox[0.98\textwidth]{\textbf{(a) Reference Images.}}
    \\
    \includegraphics[scale=0.52]{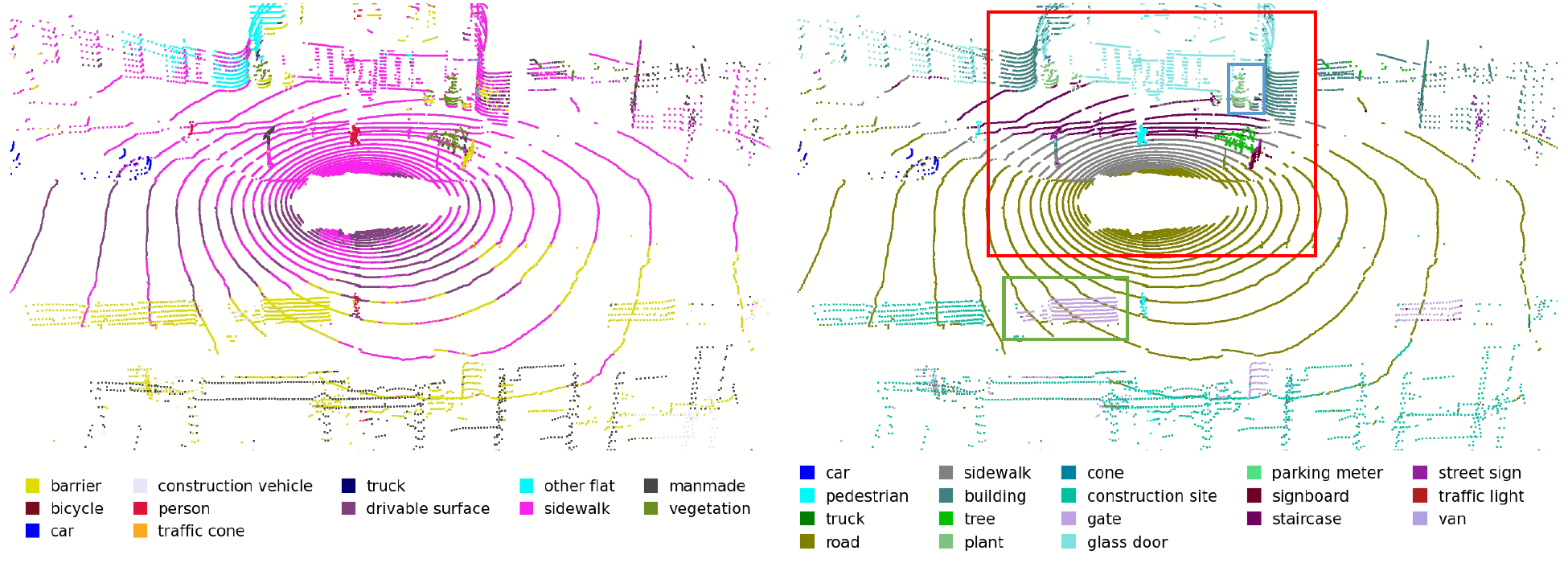}
    \\
    \vspace{-0.5em}
    \makebox[0.47\textwidth]{\textbf{(b) Pre-defined Vocabulary.}}
    \hfill
    \makebox[0.47\textwidth]{\textbf{(c) \ours (Ours)}}
    \\ 
    \caption{\textbf{Qualitative comparison between inputting pre-defined vocabulary and \ours-generated vocabulary to OpenScene~\cite{sPeng_2023_OpenScene} segmentor.} The (a) six-view images are presented for a better scene understanding. While general and ambiguous pre-defined vocabulary leads to large-area error (b). \ours segments regions with precise class names, \eg \textit{plant} (blue box), \textit{gate} (green box), \textit{road}, \textit{sidewalk}, \textit{staircase}, \textit{building} and \textit{glass door} (bottom-up in red box). These points are annotated as \textit{vegetation}, \textit{drivable surface}, \textit{sidewalk} and \textit{manmade} in the original dataset (not presented here) but are misclassified as \textit{sidewalk} and \textit{barrier} in (b).}
    \label{fig:qualitative_results}
    \vspace{-1em}
\end{figure*}

%% file: tables/quantitative.tex
\begin{table}[t]
\centering
\scriptsize
\caption{\textbf{IoU comparison on nuScenes (NUS)~\cite{nuscenes}, ScanNet (SN)~\cite{dai2017scannet} and ScanNet200 (SN200)~\cite{scannet200}.} We employ LAVE~\cite{oUlger_2023_AutoSeg} to map auto-classes from an Unknown Vocabulary (UV) to the official categories.
}
\scalebox{0.9}{\tablestyle{4pt}{1.0}
\begin{tabular}{lccccc}
\toprule
    \multirow{2}{*}{\textbf{Method}} & \multirow{2}{*}{\textbf{\makecell{Unknown\\Vocabulary}}} & \multirow{2}{*}{\textbf{Label Set}} & \multirow{2}{*}{\makecell{\textbf{NUS}\\\cite{nuscenes}}} & \multirow{2}{*}{\makecell{\textbf{SN}\\\cite{dai2017scannet}}} & \multirow{2}{*}{\makecell{\textbf{SN200}\\\cite{scannet200}}} \\
    \\
\midrule
CLIP2Scene~\cite{rChen_2023_CLIP2Scene}     & \noo & \multirow{8}{*}{Official}   & 20.8 & 25.1 & - \\
ConceptFusion~\cite{conceptfusion}          & \noo &            &- & 33.3 & 8.8\\
OpenMask3D~\cite{aTakmaz_2023_OpenMask3D}   & \noo &            & - & 34.0 & 10.3\\
HICL~\cite{himcl}                           & \noo &            & 26.8 & 33.5 & - \\
AdaCo~\cite{adaco}                          & \noo &            & \rd 31.2 & - & - \\
CNS~\cite{rChen_2023_CNS}                   & \noo &            & \nd 33.5 & 26.8 & - \\
OpenScene~\cite{sPeng_2023_OpenScene}       & \noo &            & 30.1 & \nd 47.0 & \rd 11.7 \\
Diff2Scene~\cite{diff2scene}                & \noo &            & - & \fs 48.6 & \nd 14.2 \\
3D-AVS (Ours)                               & \yes & I+L        & \fs 36.2 & \rd 40.5 & \fs 14.6 \\
\bottomrule
\end{tabular}
}
\label{tab:quantitative}
\vspace{-1em}
\end{table}

%% file: sec/6_conclusion.tex
\section{Conclusion}
\label{sec:conclusion}
In this work, we presented \ours, the first method for auto-vocabulary LiDAR point segmentation, eliminating the need for human-defined target classes.
In suboptimal image captioning conditions, our point captioner can capture missing semantics based on geometric information.
To assess the quality of the generated vocabularies in relation to segmentations, we further proposed the TPSS metric.
Our experiments show that our model's segmentations are semantically more aligned with the data than its annotations and achieves competitive masking accuracy.
We believe \ours advances scalable open-ended learning for LiDAR point segmentation without human in the loop.

%% file: sec/X0_acknowledgement.tex
\section{Acknowledgements}
\label{sec:acknowledgement}
This work was financially supported by TomTom, the University of Amsterdam and the allowance of Top consortia for Knowledge and Innovation (TKIs) from the Netherlands Ministry of Economic Affairs and Climate Policy.
Fatemeh Karimi Nejadasl was financed by the University of Amsterdam Data Science Centre.
This work used the Dutch national e-infrastructure with the support of the SURF Cooperative using grant no. EINF-7940.

%% file: sec/X_suppl.tex
\maketitlesupplementary

\section{Abstract}
This supplementary material provides additional details and analysis of our method.
Implementation details and ablation studies are presented in \cref{sec:impl_details} and \cref{sec:supp_ablations}.
We discuss the fusion of outputs from different captioners and provide qualitative comparisons under challenging lighting conditions in \cref{sec:challenging_scenes}.
The impact of the vocabulary mapper on segmentation evaluation is analyzed in \cref{sec:mapper_impact}.
Additional qualitative results are shown in \cref{sec:qua_res}.
Finally, the nomenclature used throughout the paper is summarized in \cref{sec:nomenclature}.

\section{Implementation Details} \label{sec:impl_details}
\boldparagraph{Image Captioner.}
We generate the image-based vocabulary with the xGen-MM (BLIP-3)~\cite{blip3} model using a temperature of 0.05, number of beams set to 5 and top-p set to the default value, 1. Our prompt is ``\textit{Briefly describe all objects in the \textless image\textgreater . Be concise. Only name the object names}.'', where \textless image\textgreater \ refers to the image token.

\boldparagraph{Caption Module in Point Captioner.}
We follow LidarCLIP~\cite{gHess_2024_LidarCLIP} to use the pre-trained caption model from ClipCap~\cite{clipcap} as our captioning decoder. It decodes a CLIP feature vector to one caption.

\boldparagraph{Segmenter.} 
We exploit OpenSeg~\cite{openseg} model and CLIP text encoder~\cite{aRadford_2021_CLIP} as our image encoder $h_{im}^{hr}$ and text encoder $h_{tx}$, respectively.
We employ as our point encoder OpenScene~\cite{sPeng_2023_OpenScene} with its released OpenSeg pre-trained weights on nuScenes~\cite{nuscenes} and ScanNet~\cite{dai2017scannet}, respectively.
We also follow the inference phase of OpenScene where dot production is used as similarity metric $\mathrm{SIM}$.

\boldparagraph{SMAP.}
We employ mean square error (MSE) as our loss function.
The number of views $J$ for SMAP is varying.
During training, it is set the same as the number of images per point cloud, which is six for nuScenes~\cite{nuscenes} and variable for ScanNet~\cite{dai2017scannet}.
During inference, it is set to 12 for nuScenes, indicating each point subset occupies a sector of 30 degrees.
For ScanNet, each point cloud is divided into $0.5m \times 0.5m$ squares according to their $x$ and $y$ coordinates and then each square is treated as a view.

\boldparagraph{Training.}
To train SMAP, we use Adam~\cite{AdamW} as the optimizer with an initial learning rate of $1e-5$.
The learning rate is decreased following the polynomial learning rate policy~\cite{PolyLR} with a decay of 0.9.
The SMAP has trained 20 and 10 epochs for nuScenes~\cite{nuscenes} and ScanNet~\cite{dai2017scannet}.

\section{Ablation Study} \label{sec:supp_ablations}
Ablation studies are conducted to validate our design choices and hyperparameters. 

\input{tables/ablation_image_captioner}
\boldparagraph{Image Captioner.}
\Cref{tab:image_captioner} presents the performance of 3D-AVS-Image using different image captioners.
We begin with BLIP~\cite{jLi_2022_BLIP}, but observe that it often generates low-quality nouns that are not semantically meaningful entities, such as \textit{side}, \textit{front}, or \textit{night}.
To address this limitation, we replace it with RAM~\cite{zhang2023ram} and xGen-MM~\cite{blip3}, both of which produce more precise nouns and lead to improved segmentation performance.
Moreover, RAM outputs both single and compound nouns (\eg car and asphalt road), which inspires us to enhance BLIP3 with a compound noun extraction technique that identifies consecutive nouns within captions and treats them as an individual query.
This modification yields the best overall performance.

\input{tables/ablation_point_captioner}
\input{tables/nus_details}
\boldparagraph{Point Captioner.}
\Cref{tab:point_captioner_nuscenes} shows the ablation studies of the point captioner on the nuScenes~\cite{nuscenes} dataset.
LidarCLIP generates a single caption per scene, typically covering only 2–4 common categories (\eg car and road).
In contrast, our optimal performance is achieved at $T = 12$ using a 3D relative positional encoding adapted from~\cite{xWu_2024_PTv3}, which we adopt as our final configuration.
Both LidarCLIP and our polar masking are tailored for rotating LiDAR scanners and sparse outdoor data, making them unsuitable for indoor datasets like ScanNet~\cite{dai2017scannet} and ScanNet200~\cite{scannet200}.
We instead divide the scene using vertical pillars and caption each pillar.
We find that a pillar size of 0.25 $m^2$ yields better performance (see \cref{tab:point_captioner_scannet}).
However, memory usage increases exponentially as pillar size decreases, so we set $0.5 m \times 0.5 m$ as the final resolution to avoid memory issues.

\section{Impact of Captioner Fusion.} \label{sec:challenging_scenes}

\paragraph{Quantitative Analysis.}
We report the segmentation results of 3D-AVS and its variants—using only the image captioner (3D-AVS-Image) or the point captioner (3D-AVS-Point)—on nuScenes and ScanNet in \Cref{tab:nus_details,tab:iou_scannet}.
\Cref{tab:nus_details} presents class-wise IoU on nuScenes.
The improvements are particularly notable for ambiguous categories such as \textit{drivable surface}, \textit{terrain}, and \textit{man-made}, with mIoU scores of 68.2, 41.4, and 55.4, respectively—substantially outperforming OpenScene~\cite{sPeng_2023_OpenScene}.
\Cref{tab:iou_scannet} presents the segmentation results of 3D-AVS and its variants on ScanNet.
Due to the wide variety of objects in ScanNet, 3D-AVS-LiDAR exhibits a performance drop, indicating its limited capacity for recognizing diverse object categories.

\input{tables/scannet}

\paragraph{Impact of Captioner Fusion on Challenging Scenes.} 
In \cref{sec:tpss_comparisons} and \cref{fig:tpss_barplot}, we explored the text-point similarity of generated vocabularies across various subsets of the nuScenes~\cite{nuscenes} dataset.
Our analysis indicates that in challenging conditions, such as night and rainy scenes, the point captioner outperforms the image captioner.
When the two captioners are combined, referred to as \ours, the resulting vocabularies show the strongest alignment with the data.
To illustrate this qualitatively, we present three difficult examples from the night and rainy subsets in \cref{fig:captions_challenging}.
These examples clearly show that even in these demanding scenarios, fusing the image and point captioners leads to more effective vocabulary generation, successfully identifying relevant objects in the scene. Furthermore, our method discovers additional object categories that were not originally annotated in the dataset.

\input{figures/challengingcaptions}

\section{Segmenation Performance in Relation to Vocabulary Mapper.} \label{sec:mapper_impact}

In \cref{sec:miou_comparisons} and \cref{tab:iou_scannet}, we evaluated segmentation performance using LAVE~\cite{oUlger_2023_AutoSeg} on the \nuscenes~\cite{nuscenes} and ScanNet~\cite{dai2017scannet} datasets.
To investigate the impact of different vocabulary mappers on segmentation performance, we compare three automated mappers and a manually crafted mapper on a subset of the ScanNet dataset in this section.

\paragraph{Automated Mapper.}
We evaluate segmentation performance with three automated mappers:
\begin{itemize}
    \item SentenceBERT~\cite{sbert}, which maps generated categories to target categories by measuring the similarity between two text prompts.
    \item LAVE-Llama~\cite{oUlger_2023_AutoSeg}, a LLM-based Auto-Vocabulary Evaluation (LAVE) using Llama~\cite{meta2024llama3} as the core. This method queries Llama interactively to identify the most similar target category for a given generated category.
    \item LAVE-GPT-4o, which extends LAVE by employing the more powerful GPT-4o~\cite{openai2024gpt4o} as the core language model.
\end{itemize}
Experimental results demonstrate that GPT-4o achieves the best mapping performance. Therefore, we report results using LAVE-GPT-4o as the mapper in the main text.

\paragraph{Manual Mapper.}
Given the impracticality of manual mapping for large-scale datasets, we manually mapped the automatically generated classes-125 in total-from a subset of 10 scenes in ScanNet~\cite{dai2017scannet} to the 20 original categories.
The recalculated mIoU scores, detailed in \cref{tab:handmapping}, reveal that the GPT-4o has demonstrated performance that is very close to human level on this specific task.

\input{tables/supp_other_mapping}

\input{tables/handmapping}

\input{figures/qua_res_supp}

\section{Qualitative Resutls} \label{sec:qua_res}
\Cref{fig:qualitative_results} shows the qualitative results on the ScanNet~\cite{dai2017scannet} dataset.
The complex contextual input of indoor scenarios leads to a much richer vocabulary.
Notably, the \textit{chairs} around the table in \cref{fig:qualitative_results} is misclassified as \textit{table},
while \ours successfully segments them as \textit{dining chair}.

\section{Nomenclature} \label{sec:nomenclature}
\vspace{-3em}
\input{tables/nomenclature}
\printnomenclature

%% file: tables/ablation_image_captioner.tex
\begin{table}[!t]
\caption{\textbf{Ablation study of Image Captioner on nuScenes~\cite{nuscenes} and ScanNet~\cite{dai2017scannet}.} CN (Compound Nouns) means allowing to use continuous two or more words as a query, \eg asphalt road.}
\centering
\scriptsize
\begin{tabularx}{\columnwidth}{
    p{0.15\columnwidth}
    p{0.25\columnwidth}
    >{\centering\arraybackslash}X
    >{\centering\arraybackslash}X
    >{\centering\arraybackslash}X
    >{\centering\arraybackslash}X
}
\toprule
\multirow{2}{*}{\textbf{\makecell{Ablation\\Target}}} & \multirow{2}{*}{\textbf{Setting}} & \multicolumn{2}{c}{\textbf{nuScenes}~\cite{nuscenes}} & \multicolumn{2}{c}{\textbf{ScanNet}~\cite{dai2017scannet}} \\
& & \textbf{TPSS}         & \textbf{mIoU}          & \textbf{TPSS}         & \textbf{mIoU}         \\ 
\midrule
\multirow{3}{*}{VLM}    & BLIP~\cite{jLi_2022_BLIP}       & 8.53         & 27.24         & 3.27         & 37.17        \\
                        & RAM~\cite{zhang2023ram}         & \rd 8.70         & \nd 34.14         & \rd 3.30         & \rd 38.59        \\
                        & xGen-MM~\cite{blip3}      & \nd 8.72         & \rd 33.75         & \nd 3.37         & \nd 40.27        \\
CN     & xGen-MM~\cite{blip3} + CN & \fs 8.78         & \fs 34.56         & \fs 3.49         & \fs 44.38        \\
\bottomrule
\end{tabularx}
\label{tab:image_captioner}
\end{table}

%% file: tables/ablation_point_captioner.tex
\begin{table}[t]
\caption{\textbf{Ablation study of Point Captioner on nuScenes~\cite{nuscenes}.} $T$ is a hyperparameter indicating the number of point cloud areas. LidarCLIP~\cite{gHess_2024_LidarCLIP} employs a 2D global positional encoding to generate a single global caption, whereas our method utilizes a 3D local positional encoding combined with SMAP, allowing flexible control over the number of point cloud areas to caption.}
\centering
\scalebox{0.95}{\tablestyle{6pt}{1.0}
\begin{tabular}{lcccc}
\toprule
Method                                  & \begin{tabular}[c]{@{}c@{}}$T$ \\in SMAP\end{tabular}  & \begin{tabular}[c]{@{}c@{}}Positional\\Encoding\end{tabular} & TPSS & mIoU  \\ \midrule
LidarCLIP                               & -  & 2D global                                                           & 6.25 & 20.58 \\ \midrule
Ours w/o. PE                            & 12 & \noo                                                        & 8.61 & \rd 30.89 \\ \midrule
\multirow{4}{*}{Ours}                   & 1  & \multirow{4}{*}{3D local}                                   & 6.32 & 17.94 \\
                                        & 6  &                                                             & \rd 8.66 & 29.45 \\
                                        & 12 &                                                             & \fs 8.80 & \fs 33.42 \\
                                        & 24 &                                                             & \nd 8.77 & \nd 32.96 \\
\bottomrule
\end{tabular}
}
\label{tab:point_captioner_nuscenes}
\end{table}

\begin{table}[!t]
\caption{\textbf{Ablation study of Point Captioner on ScanNet~\cite{dai2017scannet}.}}
\centering
\scriptsize
\begin{tabularx}{\columnwidth}{
    p{0.08\columnwidth}
    p{0.2\columnwidth}
    >{\centering\arraybackslash}p{0.32\columnwidth}
    >{\centering\arraybackslash}X
    >{\centering\arraybackslash}X
}
\toprule
                      & Pillar Size ($m^2$) & Num. of Pillars per Scene & TPSS & mIoU  \\
\midrule
\multirow{2}{*}{Ours} & 0.5$\times$0.5     & 87.3                & \fs 3.71 & \fs 29.25 \\
                      & 1$\times$1         & 27.6                & \nd 3.53 & \nd 22.58 \\
\bottomrule
\end{tabularx}
\label{tab:point_captioner_scannet}
\end{table}

%% file: tables/nus_details.tex
\begin{table*}[t]
\caption{\textbf{IoU comparison on nuScenes~\cite{nuscenes}.} For a quantitative comparison, we employ LAVE~\cite{oUlger_2023_AutoSeg} to map auto-classes from an Unknown Vocabulary (\textbf{UV}) to the nuScenes categories. Overall, \ours demonstrates a significant improvement over OpenScene~\cite{sPeng_2023_OpenScene}, achieving higher IoU scores on most individual labels, particularly for ambiguous classes such as \textit{drivable surface}, \textit{terrain}, and \textit{man-made}.
}
\scalebox{0.87}{\tablestyle{3.75pt}{1.2}
\newcommand{\ang}{70}
\begin{tabular}{lccccccccccccccccccc}
\toprule
\textbf{Method} & \textbf{Label Set} & \textbf{UV} & \textbf{mIoU} & \rotatebox{\ang}{barrier}  & \rotatebox{\ang}{bicycle}  & \rotatebox{\ang}{bus}  & \rotatebox{\ang}{car}  & \rotatebox{\ang}{constr. vehicle} & \rotatebox{\ang}{motorcycle}  & \rotatebox{\ang}{person}  & \rotatebox{\ang}{traffic cone} & \rotatebox{\ang}{trailer}  & \rotatebox{\ang}{truck}  & \rotatebox{\ang}{drivable surface}  & \rotatebox{\ang}{other flat} & \rotatebox{\ang}{sidewalk} & \rotatebox{\ang}{terrain}  & \rotatebox{\ang}{man-made}  & \rotatebox{\ang}{vegetation}  \\
\midrule
OpenScene~\cite{sPeng_2023_OpenScene} & Official & \noo & 30.1 & \rd 9.2  & \nd 16.3 & \nd 67.2 & \rd 70.4 & 16.4 & \nd 62.6 & \fs 47.6 & 4.0  & \fs 5.3  & 52.0 & 39.3 & 0.0 & \nd 18.1 & 0.2  & 17.5 & \rd 56.2 \\
\midrule
\multirow{3}{*}{\makecell{\ours\\(Ours)}} & Image & \yes & \nd 34.6 & \fs 13.1 & \fs 20.1 & \fs 67.6 & 65.5 & \nd 25.0 & \rd 58.9 & 2.5 & \nd 5.5 & \rd 2.8 & \fs 61.6 & \rd 52.7 & \fs 0.3 & \rd 16.8 & \nd 40.2 & \fs 55.4 & \fs 65.0 \\
                           & LiDAR & \yes & \rd 33.4 & 8.6 & 0.1 & 64.2 & \fs 72.3 & \rd 21.2 & 57.6 & \nd 44.4 & \rd 4.6 & 2.4 & \rd 55.2 & \nd 63.1 & \rd 0.1 & 12.7 & \rd 22.8 & \rd 52.1 & 53.3  \\
                           & I+L  & \yes & \fs 36.2 & \nd 12.3 & \rd 5.9 & \rd 65.1 & \nd 72.2 & \fs 25.5 & \fs 64.2 & \rd 18.0 & \fs 7.1 & \nd 4.8 & \nd 56.7 & \fs 68.2 & \nd 0.2 & \fs 20.0 & \fs 41.4 & \nd 53.5 & \nd 64.4 \\
\bottomrule
\end{tabular}
}
\label{tab:nus_details}
\end{table*}

%% file: tables/scannet.tex
\begin{table}[t]
\centering
\caption{\textbf{IoU comparison on ScanNet~\cite{dai2017scannet} validation set.}}
\scriptsize
\scalebox{0.95}{\tablestyle{11pt}{1.0}
\begin{tabular}{lccc}
\toprule
\multirow{2}{*}{\textbf{Method}} 
& \multirow{2}{*}{\textbf{\makecell{Unknown\\Vocabulary}}} 
& \multirow{2}{*}{\textbf{Label Set}} 
& \multirow{2}{*}{\textbf{mIoU}}  \\
\\
\midrule
\multirow{3}{*}{\makecell{\ours\\(Ours)}}
                       & \yes & Image                   & \nd 44.38 \\
                       & \yes & LiDAR                   &     29.25 \\
                       & \yes & I+L                     & \rd 40.51 \\
\bottomrule
\end{tabular}
}
\label{tab:iou_scannet}
\vspace{-10pt}
\end{table}

%% file: figures/challengingcaptions.tex
\begin{figure*}[t] \centering
    \includegraphics[width=\textwidth]{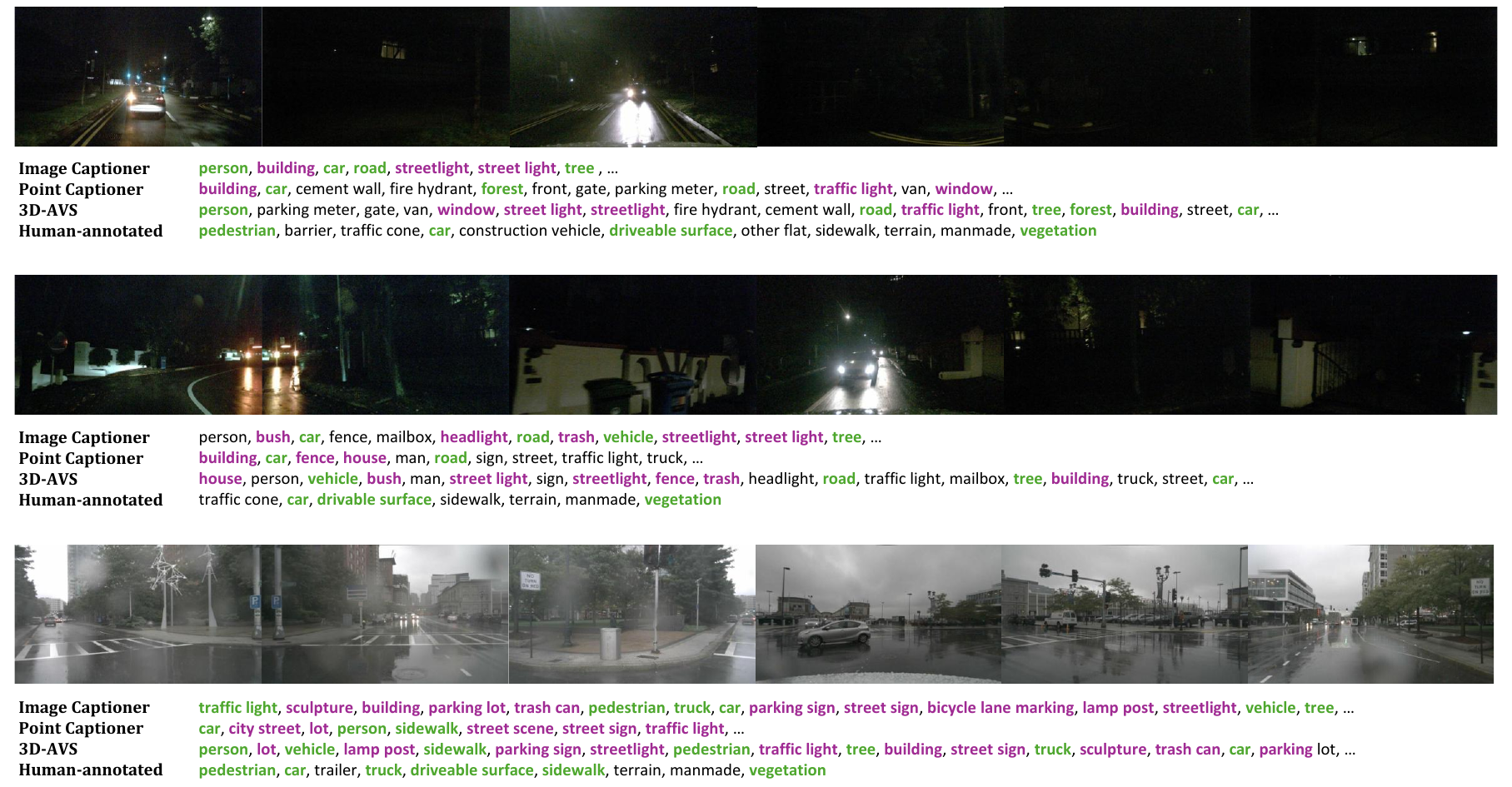}
    \\ 
    \caption{\textbf{Examples of captioners under challenging conditions.} Even in challenging weather conditions, our method is capable of generating useful descriptions of the scene, combining the strengths of both the image (when visual information is present) and the point captioner (when geometric information is present). Green classes correspond to categories that overlap with human-annotated categories provided in the dataset. Purple classes are additionally recognized by \ours which we deem plausible and useful.}
    \label{fig:captions_challenging}
\end{figure*}

%% file: tables/supp_other_mapping.tex
\begin{table}[t]
\centering
\caption{\textbf{Comparison of automated mappers on ScanNet~\cite{dai2017scannet} dataset.}}
\scriptsize
\scalebox{0.9}{\tablestyle{6pt}{1.2}
\begin{tabular}{llccc}
\toprule
\multirow{2}{*}{Method}          & \multirow{2}{*}{Label Set} & \multicolumn{2}{c}{LAVE~\cite{oUlger_2023_AutoSeg}} & \multirow{2}{*}{\makecell{Sentence\\BERT~\cite{sbert}}} \\ 
\cline{3-4}
                                 &                            & GPT-4o~\cite{openai2024gpt4o}      & Llama~\cite{meta2024llama3}      &                               \\ \midrule
\multirow{3}{*}{\makecell{3D-AVS\\(Ours)}} & Image            & \fs 44.38   & \rd 37.32  & \nd 42.60                         \\
                                 & LiDAR                      & \fs 29.25   & \nd 25.24  & \rd 23.21                         \\
                                 & Image+LiDAR    & \fs 40.51   & \rd 34.54  & \nd 39.01                         \\ \bottomrule
\end{tabular}
\label{tab:mapper_impact_scannet}
}
\vspace{-1em}
\end{table}

\begin{table}[t]
\centering
\caption{\textbf{Comparison of automated mappers on \nuscenes~\cite{nuscenes} dataset.}}
\scriptsize
\scalebox{0.9}{\tablestyle{6pt}{1.2}
\begin{tabular}{llccc}
\toprule
\multirow{2}{*}{Method}          & \multirow{2}{*}{Label Set} & \multicolumn{2}{c}{LAVE~\cite{oUlger_2023_AutoSeg}} & \multirow{2}{*}{\makecell{Sentence\\BERT~\cite{sbert}}} \\ 
\cline{3-4}
                                 &                            & GPT-4o~\cite{openai2024gpt4o}      & Llama~\cite{meta2024llama3}      &                               \\ \midrule
\multirow{3}{*}{\makecell{3D-AVS\\(Ours)}} & Image            & \fs 34.56   & \nd 33.17  & \rd 26.68                         \\
                                 & LiDAR                      & \fs 33.42   & \nd 28.92  & \rd 26.72                         \\
                                 & Image+LiDAR               & \fs 36.22   & \nd 33.68  & \rd 28.67                         \\ \bottomrule
\end{tabular}
\label{tab:mapper_impact_nuscenes}
}
\vspace{-1em}
\end{table}

%% file: tables/handmapping.tex
\begin{table}[t]
\centering
    \caption{\textbf{Mapper comparison on 10 ScanNet~\cite{dai2017scannet} validation samples.} 
            The results indicated by subscripts for LAVE ($L$) mapper with GPT-4o demonstrate performance comparable to manual mapping ($M$).}
\scriptsize
\scalebox{0.95}{\tablestyle{11.5pt}{1.0}
\begin{tabular}{llcc}
\toprule
Method & Modality & mIoU$_L$ & mIoU$_M$ \\
\midrule
\multirow{3}{*}{\makecell{3D-AVS\\(Ours)}} 
                & Image     & 30.07 & 29.81 \\
                & LiDAR        & 24.59 & 24.56 \\
                & Image+LiDAR  & 31.93 & 31.45 \\
\bottomrule
\end{tabular}
}
\label{tab:handmapping}
\end{table}

%% file: figures/qua_res_supp.tex
\begin{figure}[tbh]
\centering
    \includegraphics[width=0.93\columnwidth]{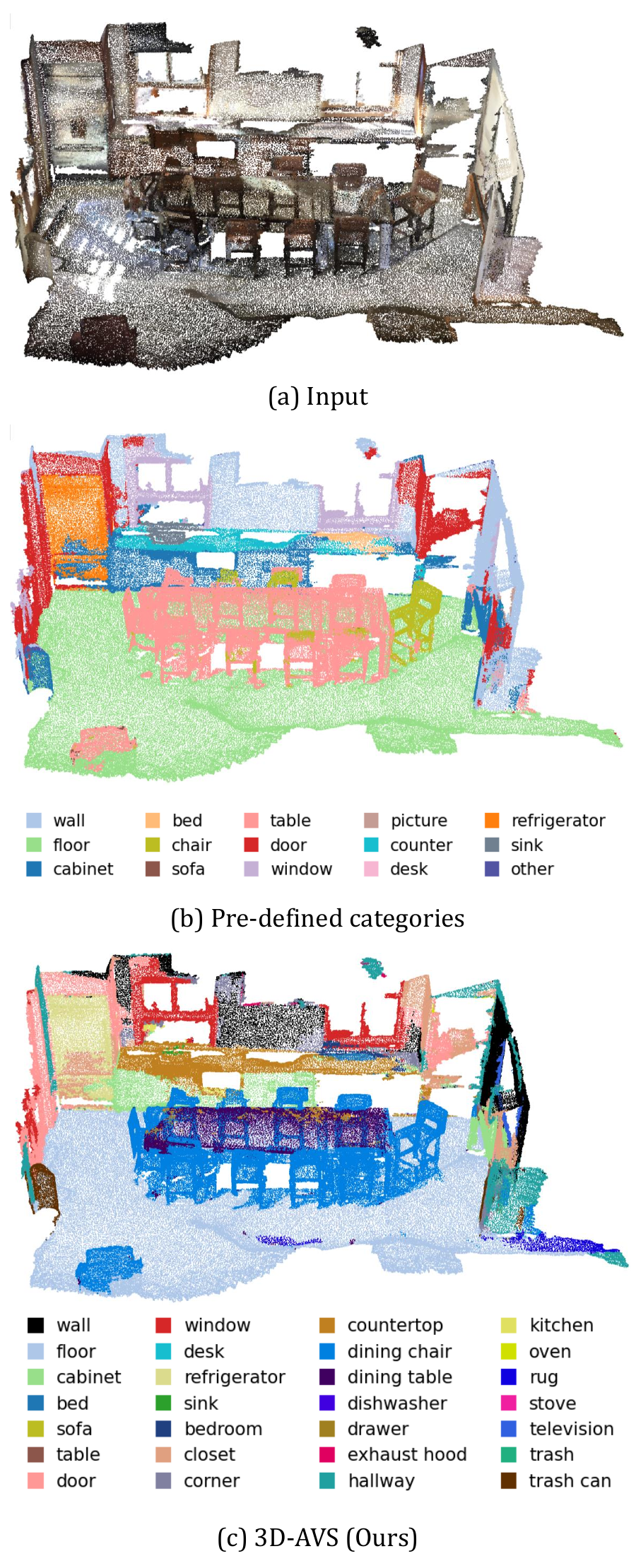}
    \\ 
    \caption{\textbf{Qualitative comparison between pre-defined categories and \ours on ScanNet Dataset~\cite{dai2017scannet}.} \ours generates much more categories than pre-defined in ScanNet. With pre-defined categories, the \textit{chair} and \textit{table} in the middle of the scene are messed up while \ours outputs a better result with generated \textit{dining chair} and \textit{dining table}.}
    \label{fig:qualitative_scannet}
\end{figure}

%% file: tables/nomenclature.tex
\nomenclcustom{M}{01}{$h_\text{tx}$}{CLIP text encoder}
\nomenclcustom{M}{02}{$h_\text{im}$}{CLIP image encoder}
\nomenclcustom{M}{03}{$h_\text{im}^\text{hr}$}{CLIP image encoder (High resolution)}
\nomenclcustom{M}{04}{$h_\text{pt}$}{CLIP point encoder}
\nomenclcustom{M}{05}{$g_\text{tx}$}{Text encoder (used in TPSS calculation)}
\nomenclcustom{M}{06}{$g_\text{pt}$}{Point encoder (used in TPSS calculation)}
\nomenclcustom{M}{11}{$\mathbf{P}$}{Point cloud, $\mathbf{P} \in \mathbb{R}^{N \times 3}$}
\nomenclcustom{M}{12}{$p_n$}{$n$-th point}
\nomenclcustom{M}{13}{$\mathbb{S}$}{Semantic space}

\nomenclcustom{M}{14}{$\mathbf{I}$}{A group of images, $\mathbf{I} \in \mathbb{R}^{K \times H \times W \times 3}$}
\nomenclcustom{M}{15}{${\dvect}_\text{im}$}{Captions from images}
\nomenclcustom{M}{16}{${\dvect}_\text{pt}$}{Captions from points}
\nomenclcustom{M}{17}{$\mathbf{L}$}{Label set, $\mathbf{L} \in \mathbb{R}^M$}
\nomenclcustom{M}{18}{$l_m$}{$m$-th label}

\nomenclcustom{M}{23}{$\mathbf{E}_\text{tx}$}{Text embeddings, $\mathbf{E} \in \mathbb{R}^{M \times C}$}
\nomenclcustom{M}{24}{$e_m$}{$m$-th text embedding}
\nomenclcustom{M}{25}{$\mathbf{F_\text{im}}$}{Image feature embeddings, $\mathbf{F_\text{im}} \in \mathbb{R}^{K \times H \times W \times C}$}
\nomenclcustom{M}{26}{$f_k$}{$k$-th image feature embedding}
\nomenclcustom{M}{27}{$\mathbf{F_\text{pt}}$}{Point feature embeddings, $\mathbf{F_\text{pt}} \in \mathbb{R}^{N \times C}$}
\nomenclcustom{M}{28}{$f_n$}{$n$-th point feature embedding}
\nomenclcustom{M}{29}{$f_n^\text{im}$}{$n$-th point feature lifted from image}
\nomenclcustom{M}{20}{$\hat{l}_n$}{Predicted label for $n$-th point}

\nomenclcustom{M}{30}{$x_n, y_n, z_n$}{Cartesian coordinates of the $n$-th point}
\nomenclcustom{M}{31}{$\rho_n$}{Radius of the $n$-th point in a polar coordinate system}
\nomenclcustom{M}{32}{$\varphi_n$}{Polar angle of the $n$-th point in a polar coordinate system}
\nomenclcustom{M}{33}{$\mathcal{B}$}{Binary masks for a point cloud, $\mathcal{B} \in \mathbb{R}^{N \times T}$ or $\mathcal{B} \in \mathbb{R}^{N \times K}$}
\nomenclcustom{M}{34}{$\mathcal{M}$}{The same as the transpose of $\mathcal{B}$. Appears in figures, $\mathcal{M} \in \mathbb{R}^{J \times N}$}
\nomenclcustom{M}{35}{$b_n^t$}{$t$-th binary mask for $n$-th point}

\nomenclcustom{M}{41}{$\mathcal{C}$}{Coordinates of a point cloud, $\mathcal{C} \in \mathbb{R}^{N \times 3}$}
\nomenclcustom{M}{42}{$\mathcal{F}$}{Features of a point cloud, $\mathcal{F} = \mathbf{F}_\text{pt} \in \mathbb{R}^{N \times C}$}
\nomenclcustom{M}{44}{$\mathcal{Q}$}{Query for MHA}
\nomenclcustom{M}{45}{$\mathcal{K}$}{Key for MHA}
\nomenclcustom{M}{46}{$\mathcal{V}$}{Value for MHA}
\nomenclcustom{M}{47}{$\mathcal{F}''$}{Output feature of SMAP, $\mathcal{F}'' \in \mathbb{R}^{J \times C}$}

\nomenclcustom{M}{51}{$\mathrm{SIM}$}{Similarity metric}
\nomenclcustom{M}{52}{$S_n$}{Similarity score for $n$-th point}

\nomenclcustom{S}{01}{$N$}{Number of points}
\nomenclcustom{S}{02}{$n$}{Index of a point}

\nomenclcustom{S}{03}{$K$}{Number of images}
\nomenclcustom{S}{04}{$k$}{Index of a image}
\nomenclcustom{S}{05}{$M$}{Number of labels}
\nomenclcustom{S}{06}{$m$}{Index of a label}
\nomenclcustom{S}{07}{$C$}{Number of channels}
\nomenclcustom{S}{08}{$T$}{Number of point cloud area}
\nomenclcustom{S}{09}{$t$}{Index of an area}
\nomenclcustom{S}{10}{$J$}{Number of binary masks}
\nomenclcustom{S}{11}{$j$}{Index of a mask}